\documentclass{bmvc2k}

\usepackage{multirow}
\usepackage[ruled,vlined]{algorithm2e}
\usepackage{bbm}
\usepackage{amsfonts}
\usepackage{booktabs}       
\definecolor{forestgreen}{rgb}{0.13, 0.55, 0.13}

\title{Inter-intra Variant Dual Representations for Self-supervised Video Recognition}

\addauthor{Lin Zhang}{https://lzhangbj.github.io}{13}
\addauthor{Qi She$\dagger$}{https://qi-she.net}{3}
\addauthor{Zhengyang Shen}{shenzhy@pku.edu.cn}{23}
\addauthor{Changhu Wang}{https://changhu.wang}{3}

\addinstitution{
School of Computer Science \\
Carnegie Mellon University \\
Pittsburgh, USA
}
\addinstitution{
School of Mathematical Science \\
Peking University \\
Beijing, China
}
\addinstitution{
ByteDance AI Lab \\
Beijing, China
}

\runninghead{ZHANG ET AL.}{INTER-INTRA DUAL VIDEO REPRESENTATION}

\begin{document}

\maketitle

\begin{abstract}
    Contrastive learning applied to self-supervised representation learning has seen a resurgence in deep models. In this paper, we find that existing contrastive learning based solutions for self-supervised video recognition focus on inter-variance encoding but ignore the intra-variance existing in clips within the same video. We thus propose to learn dual representations for each clip which (\romannumeral 1) encode intra-variance through a shuffle-rank pretext task; (\romannumeral 2) encode inter-variance through a temporal coherent contrastive loss. Experiment results show that our method plays an essential role in balancing inter and intra variances and brings consistent performance gains on multiple backbones and contrastive learning frameworks. Integrated with SimCLR and pretrained on Kinetics-400, our method achieves $\textbf{82.0\%}$ and $\textbf{51.2\%}$ downstream classification accuracy on UCF101 and HMDB51 test sets respectively and $\textbf{46.1\%}$ video retrieval accuracy on UCF101, outperforming both pretext-task based and contrastive learning based counterparts. Our code is available at \href{https://github.com/lzhangbj/DualVar}{https://github.com/lzhangbj/DualVar}.
\end{abstract}

\section{Introduction}\label{section:introduction}
Labeled data is the fundamental resource in deep learning era however is laborious to acquire. As a result, researchers resort to self-supervised learning to utilize unlabeled data. Recent rapid development of self-supervised learning \cite{chen2020simclr,caron20swav,grill2020bootstrap,He2020moco} has been largely benefited from contrastive learning \cite{wu2018instdiscrim}. With InfoNCE loss \cite{oord2019infonce}, contrastive learning tries to pull examples from the same instance (positive pairs) close while repelling those from different instances (negative pairs). This has been largely used in image tasks for its effectiveness. Meanwhile, as the most important information source in daily life, video has been actively studied towards various research directions, such as architecture design \cite{wang2021actionnet}, class incremental learning \cite{wang2020catnet} and multi-model learning \cite{wang2021teamnet}. As a result, recent works tried to transplant it into video level \cite{2021cvrl,feichtenhofer2021largescale,pan2021videomoco}, i.e. instance discrimination in the video level. Though having achieved remarkable performances, we challenge that such learning goal does not conform to the innate inter-intra variance of videos, thus is incomplete for video representation learning. Specifically, different clips sampled from different time spans of a video can exhibit different semantics. For instance, \emph{running} and \emph{jumping} are two different mini-actions though they are both sampled from a video classified as \emph{HighJump}. As contrastive learning enforces features of clips sampled from a video to always be the same, the pretrained encoder is easily overfit to the pretrained dataset. As a result, feature distributions of videos are sparse (supplementary Figure 2) and instance discrimination ability is over strong (Table \ref{table3:intraintervar}). By considering such intra-variance, pretrained encoder can be more generalizable to downstream tasks thus wins a better transferring ability. Previous works \cite{misra2016shufflelearn, opn2017, fernando2017oddout, wei2018arrowtime, xu2019vcop} have proposed to learn such temporal differences using frame/clip order verification tasks. However, the order of sub-clips is largely determined by continuity instead of semantic difference between sub-clips and is ambiguous in some repetitive actions. 
   
In this work, we delve into self-supervised video representation learning from the perspective of inter-intra variance encoding. We theoretically and experimentally find out that contrastive learning \cite{oord2019infonce} overemphasizes the learning of inter-variance but ignores intra-variance. Previous works \cite{wang20pace,chen2021rspnet} tried to separately encode inter and intra variances by appending an extra projection head to solely solve a pretext task. In contrast, we learn dual representations for each clip, which joinly encodes intra-variance between sub-clips by a shuffle-rank pretext task and inter-variance between videos by a temporal coherent contrastive loss. Besides, we adopt a ranking loss to induce a small margin between sub-clip features to reduce interference between inter and intra variance encoding. We verify its effectiveness by a series of experiments on UCF101 \cite{soomro2012ucf101} and HMDB51 \cite{Kuehne2011hmdb51} datasets and show that our method can balance inter-intra feature variances (Table \ref{table3:intraintervar}) and achieve superior performances on both finetuning and video retrieval task to state-of-the-arts.
   
In a nutshell, our contributions are $4$-fold (\romannumeral 1) We propose a shuffle-rank pretext task, which induces a small margin between intra-variant features, alleviating contradiction between inter-video and intra-video discrimination. (\romannumeral 2) We propose temporal coherent contrast on the dual representations to model inter variance learning. (\romannumeral 3) We learn joint inter-intra variant dual representations as opposed to solely inter-variant representation in contrastive learning. We also conduct a series of experiments to validate the effectiveness of our method on inter-intra variance encoding. (\romannumeral 4) The proposed method can be flexibly applied to contrastive learning frameworks, e.g. MoCo and SimCLR, with multiple spatial-temporal backbones and achieves superior performances to state-of-the-art methods. 

\section{Related work}\label{section:relatedwork}

\subsection{Self-supervised video recognition}\label{section:relatedwork_selfvideo}
We classify existing self-supervised video recognition methods into two categories based on type of the supervision signal enforced: pretext task based and contrastive learning based. 

\vspace{-1.25em}
\paragraph{Pretext tasks} Pretext task based solutions design handcrafted tasks to solve. Verifying frame and clip order \cite{misra2016shufflelearn, opn2017, fernando2017oddout, wei2018arrowtime, xu2019vcop} can provide useful order information for downstream transferring and is proved to be effective. Utilizing spatial and temporal information \cite{wang2019ststat,luo2020vcp,kim2018stpuzzle} has also achieved remarkable performances. For example, \citet{wang2019ststat} proposed to learn spatial-temporal features by designing multiple spatial-temporal statistics prediction tasks which however introduces more complexity. Recently, exploring speedness in videos have become very popular \cite{benaim2020speednet,Yao2020prp,wang20pace,jenni2020temptrans,chen2021rspnet} and also achieved state-of-the-art performance \cite{chen2021rspnet}. In this paper, we also propose a shuffle-rank pretext task to encode temporal intra-variance between sub-clips. It is essentially different from order verification in that (\romannumeral 1) We aim to learn a variety of intra-variance between clips by comparing sub-clip feature similarities instead of simply predicting clip order, which is ambiguous when clip changes are small; (\romannumeral 2) Unlike order classification, our ranking loss induces a small margin between sub-clip features, alleviating  contradiction between inter and intra variance encoding.

\vspace{-1.25em}
\paragraph{Contrastive learning} Contrastive learning tries to distinguish same instances from different ones. MoCo \cite{He2020moco} designed a negative queue to store more negatives. SimCLR \cite{chen2020simclr} proved that large batchsize is crucial to achieve superior performance. On the video level, based on MoCo, \citet{pan2021videomoco} built a temporarily decayed negative queue to model temporal variance. \citet{2021cvrl} conducted sufficient experiments to study video-level SimCLR's performance. \citet{kong2020cyclecontrast} learned feature proximity between video and frame features. \citet{feichtenhofer2021largescale} systematically analyzed four self-supervised learning frameworks on videos. However, all these contrastive learning methods aim to learn inter-video variance and intra-video invariance. Differently, \citet{tao2020iic} proposed an inter-intra contrastive learning framework by creating different positive and negative pairs but achieved little performance gains. In contrast, our work makes use of dual features to encode temporal differences between sub-clips and jointly utilized pretext tasks to achieve much better performance.

\subsection{Intra-class and inter-class variance}\label{section:relatedwork_variance}
Balancing inter and intra class variance has been a critical research field in various areas. \citet{bai2017incorporating} leveraged intra-class variance in metric learning to improve the performance of fine-grained image recognition. \citet{liu2020negative} found out that negative margins in softmax loss results in lower intra-class variance and higher inter-class variance for novel classes in few shot image classification. To alleviate long-tailed distribution, \citet{liu20longtail} proposed to increase intra-variance of tail classes by augmenting it with feature distributions of head classes. In this paper, we instead treat each video as an individual class and jointly encode instance-wise intra and inter variances in unlabeled videos.  We experimentally study the effect of our model on inter-intra variance learning in section \ref{section:effect_interintra}.

\subsection{Ranking measure}
Approximating ranking measures using functions has been studied by multiple previous works. \citet{chris2005rank} investigated using gradient descent methods to approximate ranking functions and proposed RankNet for pairwise ranking. \citet{wei2009rankmeasure} concluded that an essential loss is both an upper bound of the measure-based ranking errors and a lower bound of the loss functions. Recently, \citet{brown2020aprank} approximated ranking-based metric (Average Precision) using logistic functions and proposed Smooth-AP. \citet{varamesh2020selfrank} further applied such idea into self-supervised learning, formulating it as a ranking problem. In this work, we also use logistic function for ranking approximation. Differently, we propose to rank sub-clip features for sub-clip discrimination thus learn the intra clip variances.

\section{Preliminary}\label{section:preliminary}
In this section, we first introduce inter-intra variances in video data, and then explain the disadvantage of video contrastive learning which only encodes inter-variance. This leads to our motivation of learning inter-intra variant dual representations in section \ref{section:method}.

\subsection{Video data distribution with inter-intra variances}\label{section:preliminary_variance}
Suppose we have a collection of $N$ unlabeled videos $\{V_i\}_{i=1}^N$. Limited by memory, we sample a total of $M$ clips $\{c_i\}_{i=1}^{M}$ from videos, with $\frac{M}{N}$ clips per video. During self-supervised pretraining, a clip $c_i$ is sampled and encoded by our model $f$ into a normalized feature vector $\boldsymbol{z_i}$, i.e. $\boldsymbol{z_i} = f(c_i)$. The goal of self-supervised pretraining is to learn a good encoder $f$ that can be well transferred to downstream video action recognition. Therefore, firstly, we should distinguish different videos based on their very different contents, which is characterized as inter-variance ($\sigma_{inter}$). Secondly, semantics of clips from the same video vary a lot, e.g. \emph{running} and \emph{jumping} are two different mini-actions at different time spans of a video classified as \emph{HighJump}. Our motivation is that an encoder learning on both clip-level and sub-clip-level has a more generalized transfer ability in downstream tasks. We thus aim to produce inter and intra variant embedded features $\{\boldsymbol{z}\}$. 

\vspace{-1.25em}
\subsection{Self-supervised contrastive representation learning}\label{section:preliminary_contrast}

Contrastive learning expects clips from the same video to attract each other and repel those from different videos. Formally, the clip-feature based contrastive loss is denoted as :
\setlength{\abovedisplayskip}{0pt}
\setlength{\belowdisplayskip}{0pt}
\begin{equation}\label{equation:contrastiveloss}
    L_c = -\frac{1}{M} \sum_{i=1}^M \text{log}\frac{\text{exp}(\boldsymbol{z}_i\cdot \boldsymbol{z}_{i^+} / \tau)}{\sum_{k=1}^M \mathbbm{1}_{k\neq i} \text{exp}(\boldsymbol{z}_i\cdot \boldsymbol{z}_k/\tau)}
\end{equation}
where $\boldsymbol{z}_{i^+}$ is a positive ($+$) clip feature sampled from the same video of $\boldsymbol{z}_i$ and $\tau$ is a temperature parameter.
We can easily extend the analysis in \cite{wang20alignuniform} to find that such contrastive learning has an objective of persistently increasing $\sigma_{inter}$ and decreasing $\sigma_{intra}$, leading to insignificant intra-variance (see supplementary section 6). In this work, we propose a shuffle-rank pretext task to compensate for lack of intra-variance and a temporal coherent contrast loss between sub-clip representations to encode $\sigma_{inter}$. 

\section{Methodology}\label{section:method}

\begin{figure}[t]
    \includegraphics[width=0.8\textwidth]{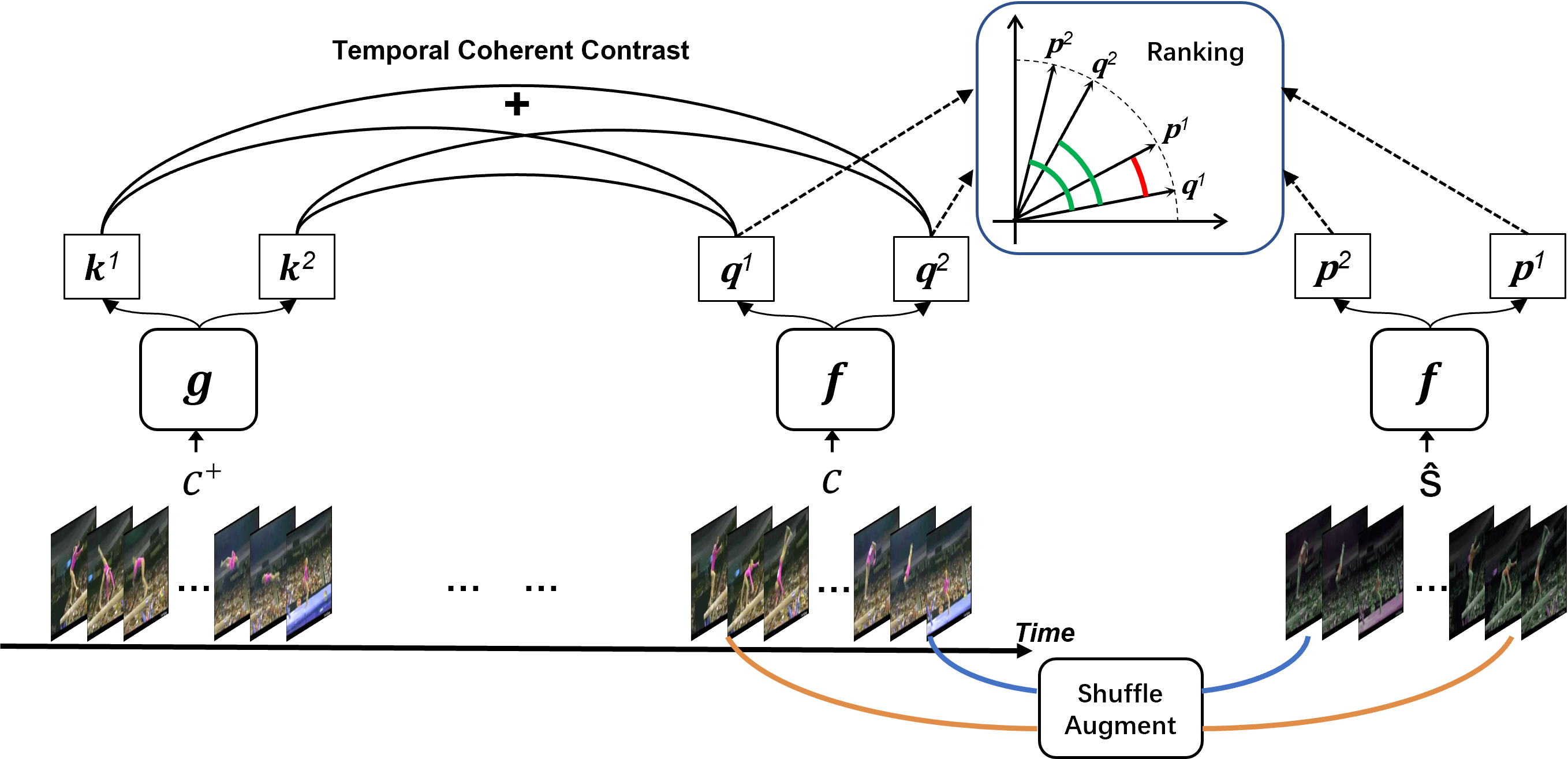}
    \centering
    \caption{Method Overview. During the intra-variance learning stage, we shuffle and augment clip $c$ into $\hat{s}$ and encode it into dual representations $\boldsymbol{p^{1}}$ and $\boldsymbol{p^{2}}$, which correspond to the unshuffled dual representations $\boldsymbol{q^{1}}$ and $\boldsymbol{q^{2}}$ respectively. If we treat $\boldsymbol{q^{1}}$ as anchor, then representation of the same sub-clip (\textcolor{red}{Red}) should be ranked higher than that of different sub-clips (\textcolor{green}{Green}). During the inter-variance learning stage (left-hand side), the temporarily variant dual representations should however keep coherence in that all dual representations of clips sampled from the same video ($c$, $c^+$) should be regarded as positive pairs (\textbf{+}) in contrastive learning. Encoder $\boldsymbol{g}$ is momentum updated by $\boldsymbol{f}$ in MoCo but equals to $\boldsymbol{f}$ in SimCLR.}
    \vspace{-1em}
    \label{figure1:overview}
\end{figure}

\subsection{Dual representations}\label{section:dualrepresentation}
In contrastive learning, an $n$-frame clip is typically encoded into a single feature for representation without considering contrast between the inner sub-clips. We instead use dual feature vectors for representing two halves of the input clip (Figure \ref{figure1:overview}). 

Formally, in addition to the original clip projector, we add another projection head, denoted as dual projection head.  A sampled clip $c$ is projected by the dual projection head into dual features $\boldsymbol{r} = (\boldsymbol{q}^1, \boldsymbol{q}^2)$. Our goal is then to jointly encode the inter and intra variances into $\boldsymbol{r}$, i.e. differences between two sub-clips of $c$ and between $c$ and other videos.

\vspace{-1em}
\subsection{Shuffle-rank}\label{section:shufflerank}
To encode differences between two sub-clips into dual representations, we propose a shuffle-rank pretext task to align raw sub-clips and learned dual features. In this section, we first describe our method, then explain the differences between our method and order prediction, which refers to simply predicting sequential order of clips and has been extensively studied before \cite{misra2016shufflelearn,kim2018stpuzzle,xu2019vcop}.

Overall, shuffle-rank consists of two stages, sub-clip shuffling and representation ranking. In the sub-clip shuffling stage, the sub-clips of an input clip $c$ is shuffled and augmented into $\hat{s}$. Both clips will then be projected into dual representations for sub-clips. In the representation ranking stage, dual features of $c$ and $\hat{s}$ will be pairwisely ranked to achieve correspondence between sub-clips and dual features through a ranking loss. As a result, the predicted dual features can genuinely reflect the intra-variance between sub-clips.

\vspace{-2em}
\paragraph{Sub-clip shuffling} We first uniformly divide clip $c$ into two sub-clips $c^1$ and $c^2$. By applying data augmentation on $c$, we get a new augmented clip $\hat{c}=(\hat{c}^1, \hat{c}^2)$, where \emph{hat} refers to augmentation. We further shuffle $\hat{c}$ to get its shuffled version $\hat{s}=(\hat{c}^2, \hat{c}^1)$. Both $c$ and $\hat{s}$ are then projected into dual representations $\boldsymbol{r}$ and $\boldsymbol{\hat{r}^s}$ through the dual projection head.
\begin{align}\label{equation:subclip_shuffle}
    c = &(c^1, c^2) \\
    \hat{c} = (\hat{c}^1, \hat{c}^2) = augment(c^1, c^2),\;&\;\;\hat{s} = (\hat{c}^2, \hat{c}^1) = shuffle(\hat{c}) \\
    \boldsymbol{r} = (\boldsymbol{q}^1, \boldsymbol{q}^2) = f(c),\;&\;\;\boldsymbol{\hat{r}^s} = (\boldsymbol{p}^2, \boldsymbol{p}^1) = f(\hat{s})
\end{align}
where $f$ is the encoding function containing a backbone and a projection layer.

\vspace{-2em}
\paragraph{Representation ranking}  Shuffle-rank alone can not guarantee the correspondence between sub-clip and dual features. Therefore, we apply a ranking measure \cite{chris2005rank,wei2009rankmeasure} to learn the subtle temporal intra-variances between sub-clips by enforcing the sub-clip feature correspondence, i.e. $\{c^1, \hat{c}^1\}\Rightarrow \{\boldsymbol{q}^1, \boldsymbol{p}^1\}$ and $\{c^2, \hat{c}^2\}\Rightarrow \{\boldsymbol{q}^2, \boldsymbol{p}^2\}$. Formally, if we regard $\boldsymbol{q}^1$ as anchor, then $\boldsymbol{p}^1$ should be ranked before both $\boldsymbol{q}^2$ and $\boldsymbol{p}^2$ while the ranking between $\boldsymbol{q}^2$ and $\boldsymbol{p}^2$ is unknown, as shown in Figure \ref{figure1:overview}. Penalties should be heavily imposed when such ranking is wrong and stay zero when the ranking is correct. However, directly applying such discrete loss would harm the stability of training. In order to have a smoother gradient backpropagation, we adopt the logistic loss function \cite{chris2005rank}. For a sub feature $\boldsymbol{x}\in\{\boldsymbol{q}^1, \boldsymbol{p}^1, \boldsymbol{q}^2, \boldsymbol{p}^2\}$, we denote its leave-self-out set of dual representations as $\boldsymbol{x}^+$ and its unpaired representation set as $\boldsymbol{x}^-$: 
\begin{align}\label{equation:representation_ranking}
    \boldsymbol{q}^{1+} = \{\boldsymbol{q}^1, \boldsymbol{p}^1\}\setminus\{\boldsymbol{q}^1\} = \{\boldsymbol{p}^1\}&,\;\;\;\boldsymbol{q}^{1-} = \{\boldsymbol{q}^2, \boldsymbol{p}^2\} \\
    \boldsymbol{q}^{2+} = \{\boldsymbol{q}^2, \boldsymbol{p}^2\}\setminus\{\boldsymbol{q}^2\} = \{\boldsymbol{p}^2\}&,\;\;\;\boldsymbol{q}^{2-} = \{\boldsymbol{q}^1, \boldsymbol{p}^1\} \\
    \boldsymbol{p}^{1+} = \{\boldsymbol{q}^1, \boldsymbol{p}^1\}\setminus\{\boldsymbol{p}^1\} = \{\boldsymbol{q}^1\}&,\;\;\;\boldsymbol{p}^{1-} = \{\boldsymbol{p}^2, \boldsymbol{q}^2\} \\
    \boldsymbol{p}^{2+} = \{\boldsymbol{q}^2, \boldsymbol{p}^2\}\setminus\{\boldsymbol{p}^2\} = \{\boldsymbol{q}^2\}&,\;\;\;\boldsymbol{p}^{2-} = \{\boldsymbol{p}^1, \boldsymbol{q}^1\}
\end{align}
Let $S$ be a function mapping two clips to their dual features set, i.e. $S(c_i, \hat{s}_i) = \{\boldsymbol{q}^1_i, \boldsymbol{q}^2_i, \boldsymbol{p}^1_i, \boldsymbol{p}^2_i \}$, then the ranking loss between \emph{unaugmented} original clips $\{c_i\}$ and their shuffled and augmented clips $\{\hat{s}_i\}$ is:
\begin{align}\label{equation:rankingloss_unaug}
    \mathcal{L}_{rank}^{unaug} = \sum_{i=1}^{M} 
    \sum_{\boldsymbol{x} \in S(c_i, \hat{s}_i)} \sum_{\boldsymbol{y} \in \boldsymbol{x}^+, \boldsymbol{z} \in \boldsymbol{x}^-}
    \text{log}(1+\text{exp}(\frac{\text{sim}(\boldsymbol{x}, \boldsymbol{z})-\text{sim}(\boldsymbol{x}, \boldsymbol{y})}{\theta}))
\end{align}
where $\theta$ is a temperature parameter. In practice, for augmentation, we also compute ranking loss between \emph{augmented} clips $\{\hat{c}_i\}$ and $\{\hat{s}_i\}$, denoted as $\mathcal{L}_{rank}^{aug}$:
\begin{align}\label{equation:rankingloss}
    \mathcal{L}_{rank}^{aug} = \sum_{i=1}^{M} 
    \sum_{\boldsymbol{x} \in S(\hat{c}_i, \hat{s}_i)} \sum_{\boldsymbol{y} \in \boldsymbol{x}^+, \boldsymbol{z} \in \boldsymbol{x}^-}
    \text{log}(1+\text{exp}(\frac{\text{sim}(\boldsymbol{x}, \boldsymbol{z})-\text{sim}(\boldsymbol{x}, \boldsymbol{y})}{\theta}))
\end{align}
Final ranking loss $\mathcal{L}_{rank} = 0.5*\mathcal{L}_{rank}^{unaug} + 0.5 * \mathcal{L}_{rank}^{aug}$. In Figure \ref{figure1:overview}, we only demonstrate the computing of $\mathcal{L}_{rank}^{unaug}$ for simplicity.

The adopted ranking loss is advantageous over order prediction in two aspects: (\romannumeral 1) Order only reflects very little information of intra-video variance, whereas in our case, by comparing the pairwise similarities between sub-clip representations, a larger variety of intra-variance can be encoded. (\romannumeral 2) Softmax cross entropy loss based order prediction induces large margin between intra-video features \cite{kobayashi2019marginsoftmax}, thus decreases the margin between inter-video features and disturbs inter-variance encoding.  Instead, ranking loss only requires a small margin between similarity of positive intra pairs ($\boldsymbol{x}$ and $y \in \boldsymbol{x}^+$) and negative intra pairs ($\boldsymbol{x}$ and $z \in \boldsymbol{x}^-$). Such a loss is also safer since sub-clip differences vary a lot from video to video, e.g. frames in a \emph{Typing} video seldom changes, exhibiting smaller intra-variance, while frames in a \emph{ClipDiving} change very fast, resulting in large intra-variance. 

In section \ref{section:compareshuffleorder}, we compare our shuffle-rank task with a common order prediction task. We also show that the temperature $\theta$ plays an important role in modeling such ranking effect and brings obvious improvement when $\theta$ is small enough.

\subsection{Temporal coherent contrastive learning}\label{section:tccontrast}
We want to further encode the inter-variance into the dual features. To do so, coherence between dual features should be maintained in that dual features from clips in the same video should be closer to each other in feature space than those from different videos, since inter-variance is much larger than intra-variance.  We thus extend clip contrast to temporal coherent contrast by using sub-clip similarity instead of clip similarity. In particular, we denote similarity between two dual representations $\boldsymbol{r}_{i}$ and $\boldsymbol{r}_{j}$ as $\text{tc-sim}(\boldsymbol{r}_i, \boldsymbol{r}_j) = \frac{1}{4} \sum_{\boldsymbol{x} \in \boldsymbol{r}_i, \boldsymbol{y} \in \boldsymbol{r}_j } \boldsymbol{x} \cdot \boldsymbol{y}$ where $\boldsymbol{r}_{i}$ and $\boldsymbol{r}_{j}$ correspond to clips $c_i$ and $c_j$ respectively. Then the temporal coherent contrastive loss is written as:
\begin{align}\label{equation:tccontrast}
    \mathcal{L}_{tc} = -\frac{1}{M} \sum_{i=1}^{M} \text{log}\frac{\text{exp}(\text{tc-sim}(\boldsymbol{r}_i, \boldsymbol{r}_{i^+})/\tau_{tc})}{\sum_{k=1}^{M} \mathbbm{1}_{[k\neq i]}\text{exp}(\text{tc-sim}(\boldsymbol{r}_i, \boldsymbol{r}_k)/\tau_{tc})}
\end{align}
where $\tau_{tc}$ is a temperature parameter and $i^+$ indexes the $i$-th clip's positive pair. Though simple, the temporal coherent contrastive learning further increases the inter-instance variances and instance discrimination ability of self-supervised learned models, and consistently improves the performance upon intra-variance encoded representations, as Table \ref{table2:generalization} shows.

Our final loss is the sum of clip contrastive loss, ranking loss and temporal coherent contrastive loss $\mathcal{L} = \mathcal{L}_c + \lambda_1 * \mathcal{L}_{rank} + \lambda_2 * \mathcal{L}_{tc}$, where $\lambda_1$ and $\lambda_2$ are hyperparameters.

\section{Experiments}
\label{experiment_section}

We conduct experiments on two contrastive learning frameworks (MoCo \cite{He2020moco}, SimCLR \cite{chen2020simclr}) and three backbones (R3D \cite{Tran2018r3d}, R(2+1)D \cite{Tran2018r3d}, S3D-G \cite{xie2018s3d}). We apply our method in pretraining stage and evaluate performance on two tasks: finetuning and video retrieval.

\subsection{Datasets}

\paragraph{Kinetics400} Kinetics400 \cite{kay2017kinetics} is a large-scale video action dataset with 400 classes and more than 400 videos for each class. All the videos are clips from Youtube and persist around 10 seconds. We are only able to obtain 218,846 videos due to invalid links. 

\vspace{-1.25em}
\paragraph{UCF101} UCF101 \cite{soomro2012ucf101} is a medium-scale human action video dataset with 13,320 videos classified into 101 classes. All the videos have a fixed frame rate of 25 FPS and a resolution of 320$\times$ 240. It provides 3 train-test splits. We use split 1 in all our experiments.

\vspace{-1.25em}
\paragraph{HMDB51} HMDB51 \cite{Kuehne2011hmdb51} is a human action video dataset with 6849 videos in 51 classes. The videos are scaled to a height of 240 pixels and 30 FPS. We use its split 1 in experiments. 

\subsection{Implementations}
We briefly introduce implementations and provide more details in supplementary section 1.

\vspace{-1.25em}
\paragraph{Self-supervised pretrain} In self-supervised pretraining stage, we randomly resize and crop clips to size of $16\times112\times112$ in a temporal consistent way with temporal stride of 4. Color jittering, horizontal flipping and gaussian blurring are applied. We pretrain the model for 200 epochs with an SGD optimizer with an initial learning rate of 0.003 and batch size of 64 on 8 Tesla V100 GPUs. We pretrain on UCF101 training split in ablation study and on large-scale Kinetics400 for performance comparison with counterparts. We set $\tau$, $\tau_{tc}$, $\theta$, $\lambda_1$, $\lambda_2$ to 0.07, 0.5, 0.05, 1.0 and 1.0, respectively.

\vspace{-1.25em}
\paragraph{Supervised finetuning} We replace the nonlinear projection head during pretraining with a classification linear layer and initialize the backbone with the pretrained weights. We finetune all layers for 150 epochs on UCF101 and HMDB51 training splits with a batchsize of 64 and learning rate of 0.05. We then test classification accuracy on test splits. 

\vspace{-1.25em}
\paragraph{Video retrieval} To evaluate the representation ability of pretrained model, we use videos in test set to retrieve videos in training set. Specifically, we average features of 10 clips uniformly sampled from each video using the pretrained backbone. We conduct video retrieval on UCF101 and calculate the top-$k$ accuracy ($k=1,5,10,20,50$).

\subsection{Ablation study}

\paragraph{Effectiveness of proposed method}\label{section:effectiveness_proposed}
We first show the effectiveness of our method by conducting experiments on both MoCo and SimCLR frameworks and three spatio-temporal backbones R3D, R(2+1)D and S3D-G. In Table \ref{table2:generalization}, consistent performance gains on multiple backbones can be observed. On SimCLR with R(2+1)D, shuffle-rank increases baseline accuracy on UCF101 and HMDB51 by $5.21\%$ and $8.27\%$ while the integrated method increases it by  $7.40\%$ and $13.59\%$ respectively.  It can be observed that performance improvements differ on different backbones, which might be due to both the internal structure of architecture and baseline performance, e.g. a strong baseline performance means smaller space for improvement. However, even on a pretty strong baseline such as MoCo with R(2+1)D backbone, our integrated method can still improve accuracy on UCF101 and HMDB51 by $0.82\%$ and $1.77\%$ respectively. Moreover, as the model is pretrained on UCF101, improvement is generally larger on HMDB51, e.g. $15.17\%$ versus $5.05\%$ with S3D-G and $13.59\%$ versus $7.40\%$ with R(2+1)D on SimCLR, verifying our model is more generalizable.

\paragraph{Effect on inter and intra variance encoding}\label{section:effect_interintra}
 To analyze the effect of our method on variance encoding, we explicitly calculate inter-intra variance of video features produced by pretrained backbone on UCF101 test set. Specifically, we uniformly sample 10 clips for each video temporarily, then calculate $\sigma_{inter}$, $\sigma_{intra}$ and instance discrimination factor $\sigma_{inter} / \sigma_{intra}$ according to the formulas defined in supplementary section 5. As shown in Table \ref{table3:intraintervar}, shuffle-rank can always increase $\sigma_{intra}$ by a large margin, e.g. $14$ times on R(2+1)D from $0.0084$ to $0.1123$. After further adding temporal coherent contrast, $\sigma_{inter}$ is increased to $0.0797$ and $\sigma_{intra}$ is decreased to $0.3798$. Our method balances the instance discrimination ability from a super high level $33.60$ to a medium value $4.77$. This general phenomenon on all three backbones (R3D, R(2+1)D, S3D-G) verifies our motivation, i.e. using shuffle-rank to encode intra-variance and temporal coherent contrast to strengthen inter-variance encoding. It also supports our statement in section \ref{section:preliminary} that encoding intra-variance can be beneficial. More experiment results on HMDB51 dataset can be see in supplementary section 3.

 \begin{table}[t]
\centering
\def\arraystretch{1.15}
\setlength\tabcolsep{2.5pt}
\begin{tabular}{ l  c c  c c  c c } 
\toprule
 \multirow{2}{*}{} & \multicolumn{2}{ c }{R3D} & \multicolumn{2}{ c }{R(2+1)D} & \multicolumn{2}{ c }{S3D-G} \\
\cline{2-7}
  & UCF101 & HMDB51 & UCF101 & HMDB51 & UCF101 & HMDB51 \\
\midrule
MoCo                & $71.72$         & $41.04$         & $77.64$         & $45.70$         & $68.41$ & $38.08$ \\
MoCo+SR        & $74.28^{\textcolor{red}{+2.56}}$ & $44.06^{\textcolor{red}{+3.02}}$ & $78.67^{\textcolor{red}{+1.03}}$ & $46.09^{\textcolor{red}{+0.39}}$ & $70.79^{\textcolor{red}{+2.38}}$ & $40.12^{\textcolor{red}{+2.04}}$\\
MoCo+SR+TC & $74.65^{\textcolor{red}{+2.93}}$ & $44.45^{\textcolor{red}{+3.41}}$ & $78.46^{\textcolor{red}{+0.82}}$ & $47.47^{\textcolor{red}{+1.77}}$ & $72.19^{\textcolor{red}{+3.78}}$ & $41.56^{\textcolor{red}{+3.48}}$ \\
\hline
SimCLR                  & $71.90$          & $39.79$         & $71.61$         & $31.78$         & $66.27$         & $20.16$ \\
SimCLR+SR          & $72.24^{\textcolor{red}{+0.34}}$  & $40.38^{\textcolor{red}{+0.59}}$ & $76.82^{\textcolor{red}{+5.21}}$ & $40.05^{\textcolor{red}{+8.27}}$ & $70.82^{\textcolor{red}{+4.55}}$ & $34.73^{\textcolor{red}{+14.57}}$ \\
SimCLR+SR+TC   & $72.69^{\textcolor{red}{+0.79}}$  & $43.01^{\textcolor{red}{+3.32}}$ & $79.01^{\textcolor{red}{+7.40}}$ & $45.37^{\textcolor{red}{+13.59}}$ & $71.32^{\textcolor{red}{+5.05}}$ & $35.33^{\textcolor{red}{+15.17}}$ \\
\bottomrule
\end{tabular}
\caption{Experiments on MoCo and SimCLR with R3D, R(2+1)D and S3D-G backbones. Models are pretrained on UCF101 train split 1. SR refers to shuffle-rank. TC refers to temporal coherent contrast. Improvement upon baseline is marked as \textcolor{red}{Red} superscripts. }
\label{table2:generalization}
\end{table}

\begin{table}[t]
\centering
\def\arraystretch{1}
\setlength\tabcolsep{0.5pt}
\begin{tabular}{ l  c c c  c c c  c c c } 
\toprule
 & \multicolumn{3}{ c }{R3D} & \multicolumn{3}{ c }{R(2+1)D} & \multicolumn{3}{ c }{S3D-G} \\ \cline{2-10}
 & inter-v. & intra-v. & discrim. & inter-v. & intra-v. & discrim. & inter-v. & intra-v. & discrim. \\
\midrule
SimCLR              & $28.8$ & $0.8$   & $34.3$ & $28.2$ & $0.8$   & $33.6$ & $73.1$ & $3.0$  & $24.7$\\
+SR        & $22.6$ & $11.3^{\textcolor{red}{\uparrow13.6\times}}$  & $2.0^{\textcolor{forestgreen}{\downarrow17.2\times}}$  & $23.0$ & $11.2^{\textcolor{red}{\uparrow13.4\times}}$  & $2.0^{\textcolor{forestgreen}{\downarrow16.8\times}}$  & $51.7$ & $11.7^{\textcolor{red}{\uparrow4.0\times}}$ & $4.4^{\textcolor{forestgreen}{\downarrow5.6\times}}$\\
+SR+TC & $42.3$ & $5.8^{\textcolor{forestgreen}{\downarrow1.9\times}}$   & $7.3^{\textcolor{red}{\uparrow3.7\times}}$  & $38.0$ & $8.0^{\textcolor{forestgreen}{\downarrow1.4\times}}$   & $4.8^{\textcolor{red}{\uparrow2.4\times}}$  & $78.7$ & $4.9^{\textcolor{forestgreen}{\downarrow2.4\times}}$  & $16.2^{\textcolor{red}{\uparrow3.7\times}}$\\
\bottomrule
\end{tabular}
\caption{Comparison of inter-instance variance, intra-instance variance and instance discrimination factor on UCF101 test set. Results are multiplied by 100 for demonstration. Each cell's increasing (\textcolor{red}{Red}) and decreasing (\textcolor{forestgreen}{Green}) times are compared to the cell above it.}
\label{table3:intraintervar}
\end{table}

\begin{figure}[t]
    \centering
    \subfigure[Graph of ranking loss. $t$ is the difference between negative and positive sub-clip pairs, i.e.  $t = \text{sim}(\boldsymbol{x}, z) - \text{sim}(\boldsymbol{x}, y)$, where $z \in \boldsymbol{x}^-$ and $y \in \boldsymbol{x}^+$.]{\includegraphics[width=0.45\textwidth]{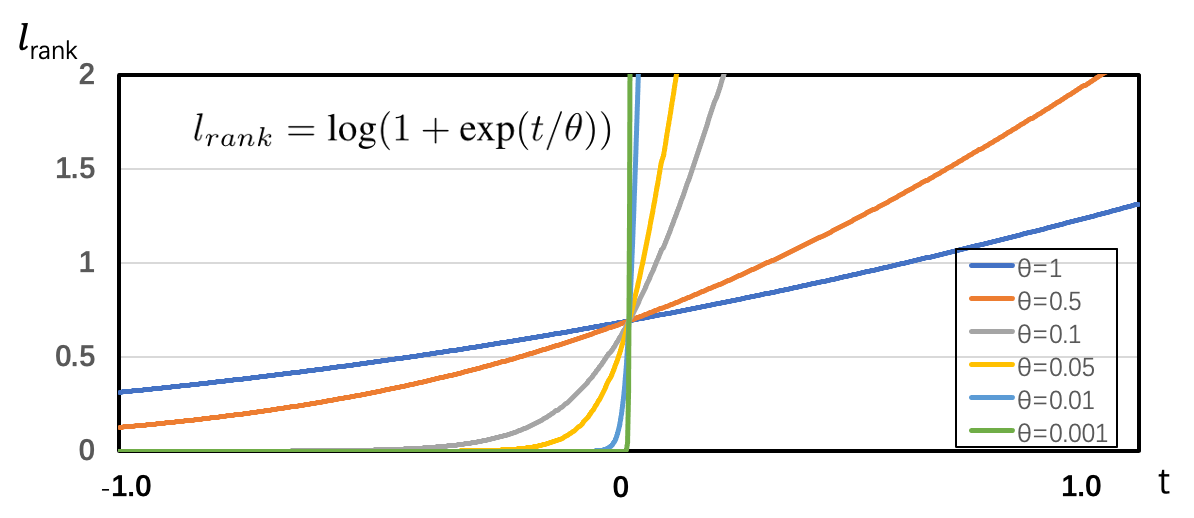}}
    \hfill
    \subfigure[Finetuning accuracies on UCF101 and HMDB51 test set.]{\includegraphics[width=0.5\textwidth]{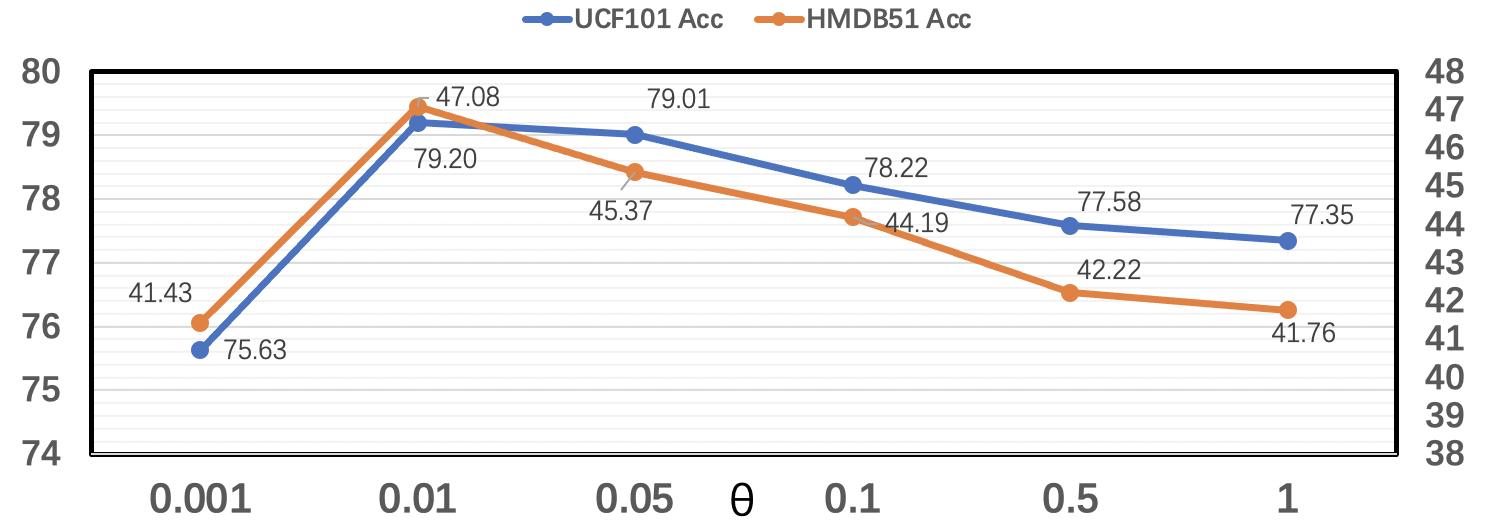}}
    \caption{Under different $\theta$ values, we (a) plot ranking loss graph (b) investigate effect of $\theta$ on downstream classification accuracy. Decreasing $\theta$ induces a smaller margin ($|t|$) under the same $l_{rank}$ and improves finetuning performance.}
    \vspace{-1em}
    \label{fig:ablationparameters}
\end{figure}
 
\vspace{-1.25em}
\paragraph{Comparison to shuffling order prediction}\label{section:compareshuffleorder}
Following our discussion in Section \ref{section:shufflerank}, we compare shuffle-rank to a non-trivial shuffling order prediction baseline. Our method is more friendly in encoding inter-intra variances by using ranking loss. Our finetuning accuracy improves upon order prediction baseline from $75.39\%$ and $32.38\%$ to $76.82\%$ and $40.05\%$ on UCF101 and HMDB51 respectively. We report details in the supplementary section 2.

\vspace{-1.25em}
\paragraph{Effect of ranking loss parameter $\theta$}\label{section:effect_theta}
Following our discussion in Section \ref{section:shufflerank}, in Figure \ref{fig:ablationparameters}, we validate our statement that inducing a smaller margin between intra positive and negative pairs brings larger benefits. In Figure \ref{fig:ablationparameters} (a), we plot ranking loss ($l_{rank}$) graph under different $\theta$. When the difference between similarities of intra negative and positive  pairs ($t = \text{sim}(\boldsymbol{x}, z) - \text{sim}(\boldsymbol{x}, y)$, where $z \in \boldsymbol{x}^-$ and $y \in \boldsymbol{x}^+ $) is zero, a fixed penalty of $log2$ is enforced as the representation is not discriminative on intra-variance. Definition of $\boldsymbol{x}^+$ and $\boldsymbol{x}^- $ is in section \ref{section:shufflerank}. As $\theta$ becomes larger, derivative at $t=0$ keeps increasing and the penalty quickly increases when the ranking measure is wrong ($t>0$) and decreases when it is correct. Besides, when the ranking is correct, penalties enforced are close to zero as long as $t$ is smaller than a margin value that is monotonically increasing with $\theta$.  As shown in Figure \ref{fig:ablationparameters} (b), as $\theta$ decreases from $1.0$ to $0.01$, model performance keeps increasing, validating our hypothesis that a small enough margin is more beneficial. However, when $\theta$ is too small as $0.001$, $l_{rank}$ is too sensitive at $t=0$, leading to unstable training. One thing need to mention here is that a smaller $\theta$ (0.01) can further increase our reported performances under $\theta=0.05$.

\begin{table}[t]
\centering
\def\arraystretch{1.0}
\setlength\tabcolsep{2pt}
\begin{tabular}{ l c c c c c c } 
\toprule
Method & Input Size & Arch & \#param. & pretrain & UCF101 & HMDB51 \\
\midrule
\textbf{Pretext Task} & & & & & \\
Shuffle\&Learn\cite{misra2016shufflelearn} & $3\times256\times256$ & AlexNet & $58.3$M & UCF101 & $50.2$ & $18.1$ \\
OPN\cite{opn2017}  & $4\times80\times80$ & VGG & $8.6$M & UCF101 & $59.8$ & $23.8$ \\
VCP\cite{luo2020vcp} & $16\times112\times112$ & R(2+1)D & $14.4$M & UCF101 & $66.3$ & $32.2$ \\
VCOP\cite{xu2019vcop} & $16\times112\times112$ & R(2+1)D & $14.4$M & UCF101 & $72.4$ & $30.9$ \\
PRP\cite{Yao2020prp} & $16\times112\times112$ & R(2+1)D & $14.4$M & UCF101 & $72.1$ & $35.0$ \\
SpeedNet\cite{benaim2020speednet} & $64\times224\times224$ & S3D-G & $9.6$M & K400 & $81.1$ & $48.8$ \\
TempTrans\cite{jenni2020temptrans} & $16\times112\times112$ & R(2+1)D-18 & $33.2$M & UCF101 & $81.6$ & $46.4$ \\
\midrule
\textbf{Contrastive}  & & & & & & \\
MemDPC\cite{Han20memdpc} & $40\times224\times224$& 3D-ResNet34 & $32.4$M & K400 & $78.1$ & $41.2$\\
VideoMoCo\cite{pan2021videomoco} & $32\times112\times112$& R(2+1)D & $14.4$M & K400 & $78.7$ & $49.2$\\
BE(MoCo)\cite{wang2021be} & $16\times112\times112$ & C3D & $27.7$M & UCF101 & $72.4$ & $42.3$ \\
IIC\cite{tao2020iic}  & $16\times112\times112$ & R3D & $14.4$M & UCF101 & $74.4$ & $38.3$ \\
\midrule
\textbf{Hybrid}  & & & & & & \\
Pace\cite{wang20pace} & $16\times112\times112$ & R(2+1)D & $14.4$M & K400 & $77.1$ & $36.6$ \\
RSPNet\cite{chen2021rspnet} & $16\times112\times112$ & R(2+1)D & $14.4$M & K400 & $81.1$ & $44.6$ \\
\midrule
Ours(MoCo)   & $16\times112\times112$ & R(2+1)D & $14.4$M & UCF101 & $78.5$ & $47.5$ \\
Ours(SimCLR) & $16\times112\times112$ & R(2+1)D & $14.4$M & UCF101 & $79.0$ & $45.4$ \\
Ours(SimCLR) & $16\times112\times112$ & R(2+1)D & $14.4$M & K400 & $\textbf{82.0}$ & $\textbf{51.2}$ \\
\bottomrule
\end{tabular}
\caption{Finetuning performance comparison.}
\label{table1:performance}
\vspace{-1.25em}
\end{table}

\begin{table}[t]
\centering
\def\arraystretch{1}
\setlength\tabcolsep{6pt}
\begin{tabular}{ l  c  c c c c c } 
\toprule
\multirow{2}{*}{} & \multirow{2}{*}{Arch} & \multicolumn{5}{c}{Top-k} \\ \cline{3-7}
& & k=1 & k=5 & k=10 & k=20 & k=50 \\
\midrule
PRP\cite{Yao2020prp}        & C3D       & $23.2$ & $38.1$ & $46.0$ & $55.7$ & $68.4$ \\
Pace\cite{wang20pace}       & R(2+1)D   & $25.6$ & $42.7$ & $51.3$ & $61.3$ & $74.0$ \\
TempTrans\cite{jenni2020temptrans} & 3D-ResNet18 & $26.1$ & $48.5$ & $59.1$ & $69.6$ & $82.8$ \\
RSPNet\cite{chen2021rspnet} & 3D-ResNet18 & $41.1$ & $59.4$ & $68.4$ & $77.8$ & $\textbf{88.7}$ \\
Ours & R(2+1)D & $\textbf{46.7}$ & $\textbf{63.1}$ & $\textbf{69.7}$ & $\textbf{78.0}$ & $87.8$ \\
\bottomrule
\end{tabular}
\caption{Video retrieval performance comparison.}
\label{table1_1:nnperformance}
\vspace{-1.25em}
\end{table}

\vspace{-1.25em}
\paragraph{Performance comparison}\label{section:performance_compare} We compare our method with previous works on both supervised finetuning and video retrieval tasks. In Table \ref{table1:performance}, we classify previous methods into 3 categories. Hybrid means combination of pretext tasks and contrastive learning. We do not compare to recent methods \cite{2021cvrl,feichtenhofer2021largescale,han20coclr} as they  use either much larger backbones and input sizes or optical flow. We outperform methods based on two mainstream pretext tasks: temporal order \cite{misra2016shufflelearn,opn2017,xu2019vcop} and pace \cite{wang20pace,benaim2020speednet,jenni2020temptrans}. SpeedNet \cite{benaim2020speednet} and TempTrans \cite{jenni2020temptrans} achieved superior performance due to large input size or backbones. MemDPC \cite{Han20memdpc} predicted future states and applies spatial-temporal contrastive loss on features however  relies on huge input size. Our model achieves higher performance than MoCo based method BE \cite{wang2021be} and VideoMoCo \cite{pan2021videomoco}. Our model surpasses RSPNet \cite{chen2021rspnet}, which is the state-of-the-art improving upon Pace \cite{wang20pace} by predicting relative speedness, by $0.9\%$ and $6.6\%$ on UCF101 and HMDB51 test set, respectively. Even pretrained on much smaller UCF101 training data, our model still exhibits excellent performance with $0.8\%$ higher HMDB51 accuracy upon Kinetics400 pretrained RSPNet. On video retrieval task in Table \ref{table1_1:nnperformance}, our method also exhibit robust performance. Our top-1 retrieval accuracy reaches $46.7\%$, improving upon RSPNet by $5.6\%$. This shows our model has a well learned discrimination ability.

\section{Conclusion}\label{section:conclusion}
In this paper, we approach self-supervised video representation learning from the perspective of inter-intra variance. We find that existing contrastive learning solution over-learns instance discrimination ability on pretrained dataset, thus has difficulty in generalization. Therefore, we propose to learn dual representations which encodes inter-intra variancesby a shuffle-rank pretext task and a temporal coherent contrast that wins a higher transferring power. It surpasses both pretext-task based and contrastive learning based counterparts on classification and video retrieval tasks on UCF101 and HMDB51 dataset.

\bibliography{egbib}
\end{document}


{\textbf{\Large Supplementary material}}

\section{Implementation details}

\paragraph{Self-supervised pretrain} In self-supervised pretraining stage, we resize each frame to 128$\times$171 and randomly crop images to size of $112\times112$ in a temporal consistent way. Each input clip has 16 frames with temporal interval of 4. Data augmentations include random color jittering, horizontal flipping and gaussian blurring. We pretrain the model for 200 epochs with an SGD optimizer with an initial learning rate of 0.003, weight decay of 1e-4 and momentum of 0.9. Learning rate is divided by 10 at epoch 120 and 160. Batch size is set to 64. We pretrain the model on 8 Tesla V100 GPUs. Pretrained weights at epoch 190 is selected for evaluation. Size of the MoCo negative queue is set to 16,384. We pretrain on UCF101 training split in all ablation studies and provide pretraining result on large-scale Kinetics400 for performance comparison with previous works. We set $\tau$, $\tau_{tc}$, $\theta$, $\lambda_1$, $\lambda_2$ to 0.07, 0.5, 0.05, 1.0 and 1.0, respectively.

\paragraph{Supervised finetuning} During supervised finetuning, we replace the nonlinear projection head in pretraining stage with a one-layer classification layer. The backbone is initialized with the pretrained weights while the classification head is randomly initialized. Similar as pretraining stage, each clip contains 16 frames sampled at a pace equals to 2. Each frame is first resized to 128$\times$171 and then randomly cropped to size of $112\times112$ in a temporal consistent way. Data augmentations include color jittering and horizontal flipping. We finetune all layers for 150 epochs using a SGD optimizer with momentum  of 0.9. Learning rate is set to 0.05 and divided by 10 at epoch 50 and 100. Batchsize is set to 64. 

\paragraph{Finetuning testing}
After finetuning, we test classification accuracy on the test splits. 10 16-frame clips are temporarily and uniformly sampled in each video with a pace equals to 2. No data augmentation is applied. Each frame is first resized to 128$\times$171 and then center cropped to size of $112\times112$. Classification probabilities are averaged among the 10 clips for each video.

\paragraph{Video retrieval} We compute average features of 10 clips uniformly sampled from each video using the pretrained backbone. Setting is the same as the \emph{Finetuning testing} stage except that we use a pace equals to 4 to be consistent with the pretraining stage. Feature vectors are computed through spatial-temporal average pooling the last layer output of the backbone and then normalized. Cosine similarity is adopted. We conduct video retrieval on UCF101 to retrieve test videos using training videos and calculate the top-$k$ accuracy ($k=1,5,10,20,50$).

\section{Comparison with shuffling order prediction}
We compare our method with shuffling order prediction baseline in Table \ref{table3:shuffleorder}. Improvement on HMDB51 dataset is much larger than that on UCF101, showing our model has a much better transferring ability across different dataset since model is pretrained on UCF101 training split.

\begin{table}[htbp]
\caption{Comparison between order prediction (OP) and our proposed shuffle-rank (SR). Models are pretrained on UCF101 training set.}
\label{table3:shuffleorder}
\centering
\def\arraystretch{1.15}
\setlength\tabcolsep{8pt}
\begin{tabular}{ l  c c} 
\toprule
 & UCF101 & HMDB51 \\
\midrule
SimCLR+OP & $75.39$ & $32.38$ \\
SimCLR+SR           & $76.82$ & $40.05$\\
\bottomrule
\end{tabular}
\end{table}

\section{More experiment results on inter-intra variance}
Continuing our analysis in ablation study section, we provide more experiment results of inter-intra variance on HMDB51 datasets for reference. As can be seen in Table \ref{table3:intraintervar}, the trend of inter/intra variance changes are consistent on both UCF101 and HMDB51.

\begin{table}[htbp]
\caption{Comparison of inter-instance variance, intra-instance variance and instance discrimination factor. Results are multiplied by 100 for demonstration. Each cell's increasing(\textcolor{red}{Red}) and decreasing(\textcolor{forestgreen}{Green}) times are compared to the cell above it.}
\label{table3:intraintervar}
\centering
\def\arraystretch{1}
\setlength\tabcolsep{0.5pt}
\begin{tabular}{ l  c c c  c c c  c c c } 
\toprule
 & \multicolumn{3}{ c }{R3D} & \multicolumn{3}{ c }{R(2+1)D} & \multicolumn{3}{ c }{S3D-G} \\ \cline{2-10}
 & inter-v. & intra-v. & discrim. & inter-v. & intra-v. & discrim. & inter-v. & intra-v. & discrim. \\
\midrule
\multicolumn{10}{c}{UCF101} \\
\hline
SimCLR              & $28.8$ & $0.8$   & $34.3$ & $28.2$ & $0.8$   & $33.6$ & $73.1$ & $3.0$  & $24.7$\\
+SR        & $22.6$ & $11.3^{\textcolor{red}{\uparrow13.6\times}}$  & $2.0^{\textcolor{forestgreen}{\downarrow17.2\times}}$  & $23.0$ & $11.2^{\textcolor{red}{\uparrow13.4\times}}$  & $2.0^{\textcolor{forestgreen}{\downarrow16.8\times}}$  & $51.7$ & $11.7^{\textcolor{red}{\uparrow4.0\times}}$ & $4.4^{\textcolor{forestgreen}{\downarrow5.6\times}}$\\
+SR+TC & $42.3$ & $5.8^{\textcolor{forestgreen}{\downarrow1.9\times}}$   & $7.3^{\textcolor{red}{\uparrow3.7\times}}$  & $38.0$ & $8.0^{\textcolor{forestgreen}{\downarrow1.4\times}}$   & $4.8^{\textcolor{red}{\uparrow2.4\times}}$  & $78.7$ & $4.9^{\textcolor{forestgreen}{\downarrow2.4\times}}$  & $16.2^{\textcolor{red}{\uparrow3.7\times}}$\\
\midrule
\multicolumn{10}{c}{HMDB51} \\
\hline
SimCLR              & $26.6$ & $0.3$ & $85.8$ & $25.8$ & $0.3$ & $83.3$ & $71.1$ & $1.2$ & $58.3$\\
+SR        & $33.7$ & $5.8^{\textcolor{red}{\uparrow18.7\times}}$ & $5.8^{\textcolor{forestgreen}{\downarrow14.8\times}}$  & $32.2$ & $5.1^{\textcolor{red}{\uparrow16.4\times}}$ & $6.4^{\textcolor{forestgreen}{\downarrow13.0\times}}$  & $61.8$ & $4.8^{\textcolor{red}{\uparrow3.9\times}}$ & $12.9^{\textcolor{forestgreen}{\downarrow4.5\times}}$\\
+SR+TC & $43.0$ & $2.9^{\textcolor{forestgreen}{\downarrow2.0\times}}$ & $14.8^{\textcolor{red}{\uparrow2.6\times}}$ & $40.0$ & $3.8^{\textcolor{forestgreen}{\downarrow1.3\times}}$ & $10.6^{\textcolor{red}{\uparrow1.7\times}}$ & $73.3$ & $2.5^{\textcolor{forestgreen}{\downarrow1.9\times}}$ & $29.0^{\textcolor{red}{\uparrow2.2\times}}$\\
\bottomrule
\end{tabular}
\end{table}

\section{Result of different number of segments}
We ablated the number of segments in Table \ref{table_x:n_segment}. Model is equipped with R(2+1)D backbone and pretrained on UCF101 training split. As input clip length is the same, increasing number of segments also increases the similarity between temporarily neighboring segments, which largely increases the difficulty for shuffle-rank task. We find increase segment number from 2 to 4 does not improve the performance much, but 8 segments make the loss not converging, which is caused by the difficulty of the task.

\begin{table}[htbp]
\centering
\def\arraystretch{1}
\setlength\tabcolsep{6pt}
\begin{tabular}{ c  c  c } 
\toprule
Number of Segment & UCF101 & HMDB51 \\
\midrule
2 & $79.01$ & $45.37$ \\
4 & $77.32$ & $46.05$ \\
8 & $null$ & $null$ \\
\bottomrule
\end{tabular}
\caption{Finetune task result on different number of segments with R(2+1)D backbone pretrained on UCF101 training split. $null$ means loss can not converge in training time.}
\label{table_x:n_segment}
\vspace{-1.25em}
\end{table}

\section{Per-class improvement result}
\begin{figure}[t]
    \includegraphics[width=0.8\textwidth]{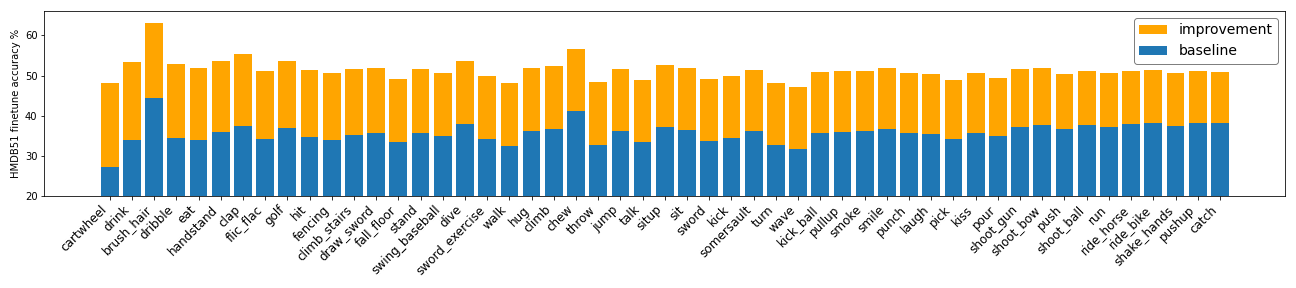}
    \centering
    \caption{Per-class improvement on HMDB51 downstream supervised finetuning task with SimCLR baseline and R(2+1)D backbone. Model is unsupervisedly pretrained on UCF101 training split and $\theta$ is chosen as 0.01. The category labels along the horizontal coordinate follows decreasing order of absolute improvement magnitudes upon the baseline (height of orange bars).}
    \vspace{-1em}
    \label{figure3:per_class_imprv}
\end{figure}

\paragraph{Per-class Analysis}\label{section:per_class_improvement} We provide per-class accuracy improvement result of our method upon baseline SimCLR in  Figure \ref{figure3:per_class_imprv}. Remind that in HMDB51 dataset, highly-improved categories such as cartwheel, drink, eat and handstand usually contain decomposable unrepetitive actions with large motion changes, while categories like running, riding horses, shake-hands consist of many repetitive actions thus are less improved by our method. All actions have been improved. These results supported our motivation that representing videos by sub-features are useful.

\section{Backbone architecture}
We provide architectures of R3D and R(2+1)D in Table \ref{table:arch}. As the structure of S3D-G is too complicated to be put in a table, we ask the reader to refer to \cite{xie2018s3d} for more  details. We use the same implementation in our code.
\begin{table}[htbp]
\caption{architecture of R3D and R(2+1)D we used in experiments.}
\label{table:arch}
\centering
\def\arraystretch{1.5}
\setlength\tabcolsep{8pt}
\begin{tabular}{ c | c  | c | c } 
\hline
layer name & output size & R3D & R(2+1)D \\\hline
conv1 & $L\times56\times56$ & \multicolumn{2}{c}{$3\times3\times3$, 64, stride $1\times2\times2$} \\\hline
conv2\_x & $L\times56\times56$ & \begin{bmatrix}3\times3\times3,64\\3\times3\times3, 64\end{bmatrix} $\times$1 & \begin{bmatrix}1\times3\times3,64\\3\times1\times1, 64\end{bmatrix} $\times$2 \\\hline
conv3\_x & $\frac{L}{2}\times28\times28$ & \begin{bmatrix}3\times3\times3,128\\3\times3\times3, 128\end{bmatrix} $\times$1 & \begin{bmatrix}1\times3\times3,128\\3\times1\times1, 128\end{bmatrix} $\times$2\\\hline
conv4\_x & $\frac{L}{4}\times14\times14$ & \begin{bmatrix}3\times3\times3,256\\3\times3\times3, 256\end{bmatrix} $\times$1 & \begin{bmatrix}1\times3\times3,256\\3\times1\times1, 256\end{bmatrix} $\times$2\\\hline
conv5\_x & $\frac{L}{8}\times7\times7$ & \begin{bmatrix}3\times3\times3,512\\3\times3\times3, 512\end{bmatrix} $\times$1  & \begin{bmatrix}1\times3\times3,512\\3\times1\times1, 512\end{bmatrix} $\times$2\\\hline
        & $1\times1\times1$ & \multicolumn{2}{ c }{spatial-temporal global average pooling} \\\hline
\end{tabular}
\end{table}

\section{Measure inter and intra variances}
We provide our mathematical formulations of inter-variance and intra-variance for calculating inter and intra variances. Formally,  let $\mu_k$ represent the mean embedded feature of the $k$-th video.  We use $S_{k}$ to represent the hierarchical relation between clips $\{c_i\}_{i=1}^M$ and videos $\{V_i\}_{i=1}^N$, i.e. $S_{k} = \{i | c_i \in V_k\}$. We define intra-variance as the average variances within clip features sampled from the same video, and inter-variance as the average pairwise distance between videos: 
\begin{align}
\mu_k & = \frac{1}{S_k} \sum_{i\in S_k} \boldsymbol{z}_i \\\sigma_{intra} & = \frac{1}{N} \sum_{k=1}^{N} \frac{1}{|S_{k}|} \sum_{i\in{S_{k}}} \|\boldsymbol{z}_i - \mu_k\|_2^2 \\\sigma_{inter} & = \frac{1}{N(N-1)} \sum_{i=1}^{N-1} \sum_{j=i+1}^{N} {\| \mu_i -\mu_j \|}_2^2
\end{align}
where $\boldsymbol{z}$ is the encoded feature vector as introduced in the paper. This formulation is adapted from \citet{liu2020negative}.

\section{Effect of contrastive loss on inter and intra variance}
We provide theoretical analysis to our statement in preliminary section that contrastive loss suppresses the learning of $\sigma_{inter}$ and encourages learning $\sigma_{inter}$. Our analysis follows discussion of \citet{wang20alignuniform}. Following our definition in preliminary section, clip contrastive loss can be written in the form of $\mathcal{L}_c = -\frac{1}{M}\sum_{i=1}^{M} \text{log}\frac{\text{exp}(\frac{\boldsymbol{z}_i\cdot \boldsymbol{z}_{i^+}}{\tau})}{\sum_{k=1}^{M} \bold{\mathbbm{1}_{[k\neq i]}}\text{exp}(\frac{\boldsymbol{z}_i\cdot \boldsymbol{z}_k}{\tau})}$, which can be transformed into:
\begin{align}\label{equation:contrastivetransfrom}
    \mathcal{L}_c &= -\frac{1}{M}\sum_{i=1}^{M} \frac{\boldsymbol{z}_i\cdot \boldsymbol{z}_{i^+}}{\tau} + \frac{1}{M}\sum_{i=1}^{M} \left[ \text{log}\left( \text{exp}(\frac{\boldsymbol{z}_i\cdot \boldsymbol{z}_{i^+}}{\tau}) + \sum_{k=1, k\not\in \{ i, i^+\} }^{M} \text{exp}(\frac{\boldsymbol{z}_i\cdot \boldsymbol{z}_k}{\tau}) \right)  \right]
\end{align}
where $M$ is the number of sampled clips, $\boldsymbol{z}$ is normalized encoded clip feature vector and $\tau$ is the temperature parameter. As explained in \cite{wang20alignuniform}, since $\sum_{k=1, k\not\in \{ i, i^+\} }^{M} \text{exp}(\boldsymbol{z}_i\cdot \boldsymbol{z}_k/\tau)$ is always positive and bounded below, the loss favors smaller$-\frac{1}{M}\sum_{i=1}^{M} \boldsymbol{z}_i\cdot \boldsymbol{z}_{i^+}/\tau$, which is equivalent to decreasing $\sigma_{intra}$ as the sample size $M$ goes to infinity. On the other hand, when fixing $\sigma_{intra}$, minimizing the loss is equivalent to decreasing 
\begin{align}\label{equation:uniformitem}
     \frac{1}{M}\sum_{i=1}^{M} \left[ \text{log} ( t_{i} + \sum_{k=1, k\not\in \{ i, i^+\} }^{M} \text{exp}(\boldsymbol{z}_i\cdot \boldsymbol{z}_k/\tau) ) \right]
\end{align}
where $t_i = \text{exp}(\boldsymbol{z}_i\cdot \boldsymbol{z}_{i^+}/\tau) $ is a random variable when $\sigma_{intra}$ is fixed. Hence, equation \ref{equation:uniformitem} measures similarity between features from different instances. Minimizing equation \ref{equation:uniformitem} is thus equivalent to minimizing inter-variance $\sigma_{inter}$. Therefore, contrastive loss encourages decreasing intra-variance and increasing inter-variance.

\section{Training algorithm}



\subsection{Training SimCLR}

\begin{algorithm}[H]
\setstretch{1.2}
\SetAlgoLined
\SetNoFillComment
 \textbf{Input}: batch size $N$, backbone $f$, clip projector $f_c$, dual projector $f_r$, videos $V$\;
 \While{training not converge}{
    sample minibatch $\{v_i\}_{i=1}^{N}$ from $V$\;
    \For{all $i \in \{1,2,...N\}$}{
        sample and augment two clips $c_i^1$ and $c_i^2$ from $v_i$ \;
        $\boldsymbol{h}_i^1 = f(c_i^1)$ \;
        $\boldsymbol{h}_i^2 = f(c_i^2)$ \;
        \tcc{clip projection}
        $\boldsymbol{z}_{2i-1} = f_c(\boldsymbol{h}_i^1)$ \;
        $\boldsymbol{z}_{2i} = f_c(\boldsymbol{h}_i^2)$ \;
        \tcc{dual projection}
        $\boldsymbol{r}_{2i-1} = f_r(\boldsymbol{h}_i^1)$ \;
        $\boldsymbol{r}_{2i} = f_r(\boldsymbol{h}_i^2)$ \;
        \tcc{shuffling and dual projection}
        augment $c_i^1$ into $s_i$ and randomly shuffle $s_i$ into $\hat{s}_i$ \;
        $\boldsymbol{q}_i = f_r(f(s_i))$ \;
        $\boldsymbol{p}_i = f_r(f(\hat{s}_i))$ \;
    }
    calculate $\mathcal{L}_c$ on $\{\boldsymbol{z}\}_{i=1}^{2N}$\;
    calculate $\mathcal{L}_{tc}$ on $\{\boldsymbol{r}\}_{i=1}^{2N}$\;
    calculate $\mathcal{L}_{rank}^{unaug}$ on $\{\boldsymbol{r}_{2i-1}, \boldsymbol{p}_i \}_{i=1}^{N}$ and $\mathcal{L}_{rank}^{aug}$ on $\{\boldsymbol{q}_i, \boldsymbol{p}_i\}_{i=1}^{N}$\;
    update $f$, $f_c$ and $f_r$ using stochastic gradient descent\;
 }
 \caption{Training SimCLR}
\end{algorithm}

\subsection{Training MoCo}
Unlike SimCLR, negative examples in MoCo come from the dictionary $\{\boldsymbol{k}_i\}_{i=1}^{m}$ while positive example is the sampled augmented clip $\boldsymbol{k}^+$. $m$ is set to 16384. So contrastive loss is denoted as :

\begin{align}\label{equation:moco_clipcontrast}
    \mathcal{L}_c = -\frac{1}{N} \sum_{i=1}^{N} \frac{\text{exp}(\boldsymbol{z}_i \cdot \boldsymbol{k}_i^+ /\tau)}{\text{exp}(\boldsymbol{z}_i \cdot \boldsymbol{k}_i^+ /\tau) + \sum_{j=1}^{m} \text{exp}(\boldsymbol{z}_i \cdot \boldsymbol{k}_j /\tau)}
\end{align}

As for temporal coherent contrast, an extra dictionary for dual representations $\{\boldsymbol{d}_i\}_{i=1}^m$ is maintained. Dual representations are denoted as $\boldsymbol{r}$ instead of $\boldsymbol{z}$. Replacing dot product similarity, dictionary $\boldsymbol{k}$, projected feature $\boldsymbol{z}$ with tc-sim, $\boldsymbol{d}$ and $\boldsymbol{r}$ in equation \ref{equation:moco_clipcontrast} leads us to MoCo's temporal coherent contrast loss:

\begin{align}\label{equation:moco_tccontrast}
    \mathcal{L}_{tc} = -\frac{1}{N} \sum_{i=1}^{N} \frac{\text{exp}( \text{tc-sim}(\boldsymbol{r}_i, \boldsymbol{d}_i^+ )/\tau)}{\text{exp}( \text{tc-sim}(\boldsymbol{r}_i, \boldsymbol{d}_i^+ )/\tau) + \sum_{j=1}^{m} \text{exp}( \text{tc-sim}(\boldsymbol{r}_i, \boldsymbol{d}_j) /\tau)}
\end{align}

We provide training algorithm on MoCo here.

\begin{algorithm}[H]
\setstretch{1.2}
\SetAlgoLined
\SetNoFillComment
 \textbf{Input}: batch size $N$, backbone $f_q$ $f_k$, clip projectors $g_q$ $g_k$, dual projectors $h_q$ $h_k$, videos $V$, clip feature dictionary $\{\boldsymbol{k}_i\}_{i=1}^m$ and dual feature dictionary $\{\boldsymbol{d}_i\}_{i=1}^m$\;
 \While{training not converge}{
    sample minibatch $\{v_k\}_{k=1}^{N}$ from $V$\;
    \For{all $i \in \{1,2,...N\}$}{
        sample and augment two clips $c_i^1$ and $c_i^2$ from $v_i$, $c_i^2$ is treated as positive pair of $c_i^1$ \;
        $\boldsymbol{b}_i^1 = f_q(c_i^1)$ \;
        $\boldsymbol{b}_i^2 = f_k(c_i^2)$ \;
        \tcc{clip projection}
        $\boldsymbol{z}_i = g_q(\boldsymbol{b}_i^1)$ \;
        $\boldsymbol{k}_i^+ = g_k(\boldsymbol{b}_i^2)$ \;
        $\boldsymbol{k}_i^+ = \boldsymbol{k}_i^+.detach()$ \tcp*{stop gradient}
        \tcc{dual projection}
        $\boldsymbol{r}_i = h_q(\boldsymbol{b}_i^1)$ \;
        $\boldsymbol{d}_i^+ = h_k(\boldsymbol{b}_i^2)$ \;
        $\boldsymbol{d}_i^+ = \boldsymbol{d}_i^+.detach()$ \tcp*{stop gradient}
        \tcc{shuffling and dual projection}
        augment $c_i^1$ into $s_i$ and randomly shuffle $s_i$ into $\hat{s}_i$ \;
        $\boldsymbol{q}_i = h_q(f_q(s_i))$ \;
        $\boldsymbol{p}_i = h_q(f_q(\hat{s}_i))$ \;
    }
    calculate $\mathcal{L}_c$ on positive pairs $\{\boldsymbol{z}_i, \boldsymbol{k}_i^+\}_{i=1}^{N}$ and negative queue $\{\boldsymbol{k}_i\}_{i=1}^m$ \;
    calculate $\mathcal{L}_{tc}$ on positive pairs  $\{\boldsymbol{r}_i, \boldsymbol{d}_i^+\}_{i=1}^{N}$ and negative queue $\{\boldsymbol{d}_i\}_{i=1}^m$ \;
    calculate $\mathcal{L}_{rank}^{unaug}$ on $\{\boldsymbol{r}_i, \boldsymbol{p}_i \}_{i=1}^{N}$ and $\mathcal{L}_{rank}^{aug}$ on $\{\boldsymbol{q}_i, \boldsymbol{p}_i\}_{i=1}^{N}$ \;
    momentum update $f_k$, $g_k$ and $h_k$ by $f_q$, $g_q$ and $h_q$ respectively\;
    update $f_q$, $g_q$ and $h_q$ using stochastic gradient descent\;
    $enqueue(\{\boldsymbol{k}_i\}_{i=1}^m, \{\boldsymbol{k}_i^+\}_{i=1}^{N})$\;
    $dequeue(\{\boldsymbol{k}_i\}_{i=1}^m)$\;
    $enqueue(\{\boldsymbol{d}_i\}_{i=1}^m, \{\boldsymbol{d}_i^+\}_{i=1}^{N})$\;
    $dequeue(\{\boldsymbol{d}_i\}_{i=1}^m)$\;
 }
 \caption{Training MoCo}
\end{algorithm}


\section{Feature distribution comparison}
We visualize the feature distribution encoded by baseline SimCLR and our pretrained models on UCF101 test set. As shown, feature distribution of baseline SimCLR is very sparse, with large margin between different classes. With our shuffle-rank pretext task, intra-variance becomes larger, resulting in a denser distribution. Further equipping the model with temporal coherent contrastive loss makes the clustering borders clearer. This is consistent with our inter-intra variance motivation.

\section{Attention visualization}
To  qualitatively measure the learned features, we draw heatmaps output from last layer of pretrained models. Specifically, we apply average pooling on last layer feature maps along both feature channel and temporal channel, i.e. from a size of $( C, T, H, W )$ into $(H, W)$, to obtain a heatmap. Such heatmap is then added to each frame of the input clip for visualization. As shown in Figure \ref{fig:heatmap}, baseline SimCLR can be easily distracted by surrounding backgrounds, while our proposed method always focus on areas with motion semantics.

\begin{figure}[h]
    \centering
    \subfigure[SimCLR]{\includegraphics[width=0.3\textwidth]{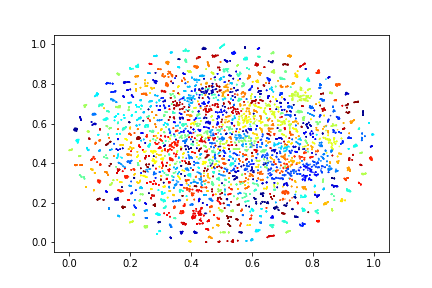}} 
    \subfigure[SimCLR+SR]{\includegraphics[width=0.3\textwidth]{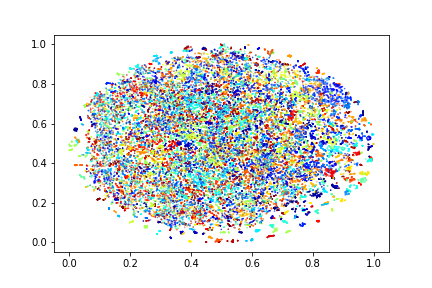}} 
    \subfigure[SimCLR+SR+TC]{\includegraphics[width=0.3\textwidth]{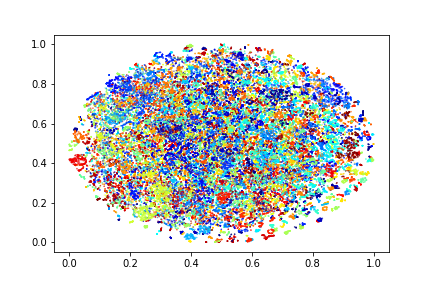}}
    \caption{t-SNE visualization of features encoded by pretrained models. Models are pretrained on UCF101 train split and features are computed on UCF101 test split.}
\end{figure}
\begin{figure*}[h]
\centering
\def\arraystretch{2.5}
\begin{tabular}{ l c @{\hspace{1.5\tabcolsep}} c @{\hspace{1.5\tabcolsep}}}
Raw Clip &
\multirow{4}{*}{\includegraphics[width=5cm,valign=c]{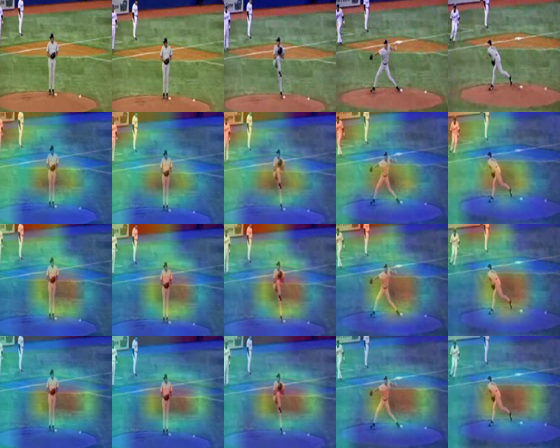}} &
\multirow{4}{*}{\includegraphics[width=5cm,valign=c]{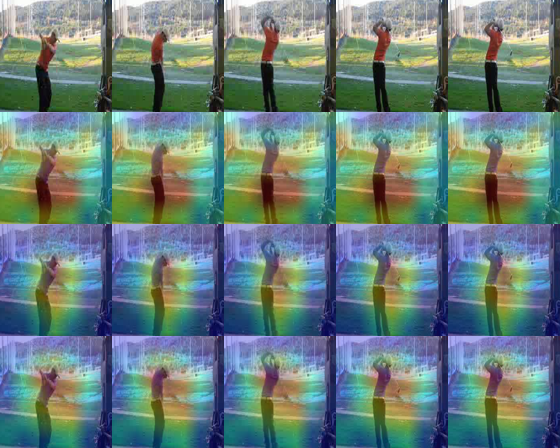}} \\
SimCLR & & \\
SimCLR+SR & & \\
SimCLR+SR+TC & & \\[2ex]
Raw Clip &
\multirow{4}{*}{\includegraphics[width=5cm,valign=c]{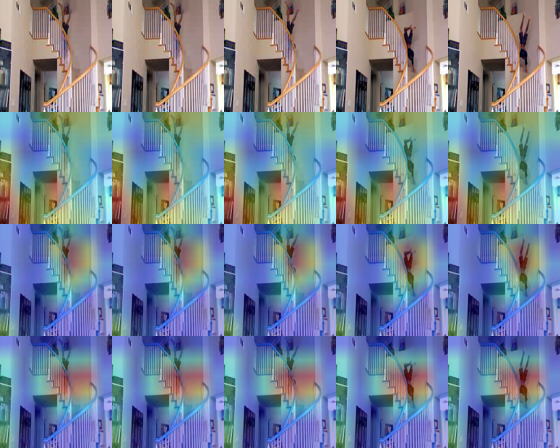}} &
\multirow{4}{*}{\includegraphics[width=5cm,valign=c]{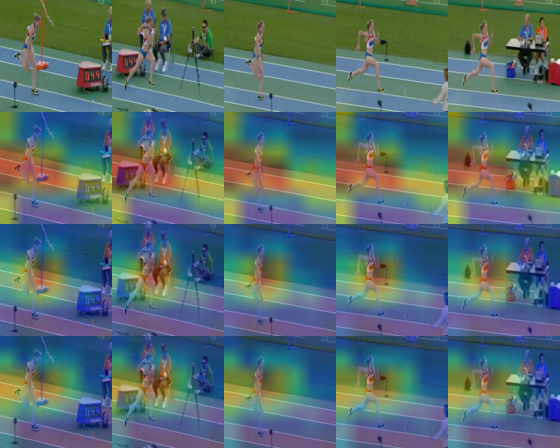}} \\
SimCLR & & \\
SimCLR+SR & & \\
SimCLR+SR+TC & & \\[2ex]
Raw Clip &
\multirow{4}{*}{\includegraphics[width=5cm,valign=c]{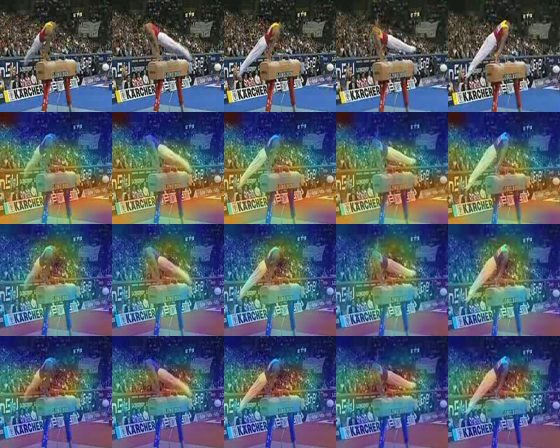}} &
\multirow{4}{*}{\includegraphics[width=5cm,valign=c]{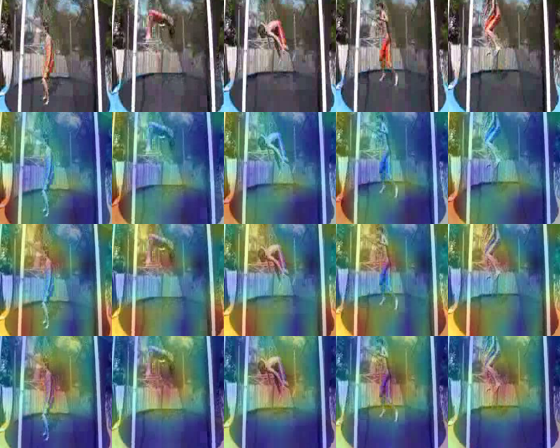}} \\
SimCLR & & \\
SimCLR+SR & & \\
SimCLR+SR+TC & & \\[2ex]
\end{tabular}
\caption{Heatmap visualization. The warmer the color is, the larger the response is. Zoom in for visualization.}
\label{fig:heatmap}
\end{figure*}

\section{Video retrieval visualization}
\begin{figure}[h]
\centering
\begin{tabular}{ l c @{\hspace{1.5\tabcolsep}} c @{\hspace{0.5\tabcolsep}} c @{\hspace{0.5\tabcolsep}} c}
\begin{tabular}{c} ApplyEyeMakeup\end{tabular} &
\includegraphics[width=2.5cm]{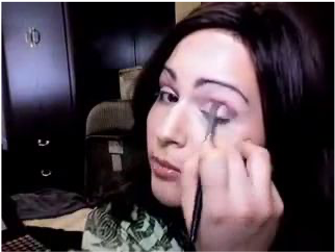}&
\includegraphics[width=2.5cm]{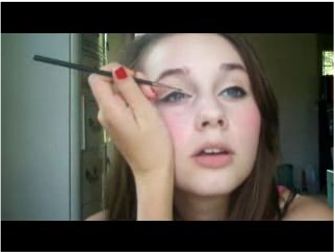}&
\includegraphics[width=2.5cm]{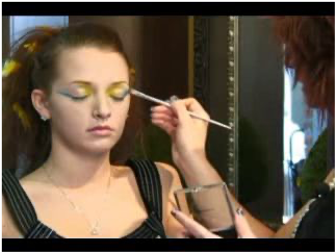}&
\includegraphics[width=2.5cm]{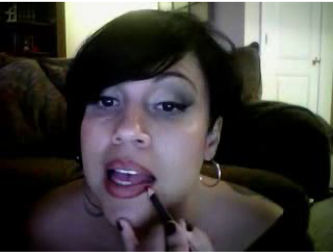} \\
BabyCrawling & 
\includegraphics[width=2.5cm]{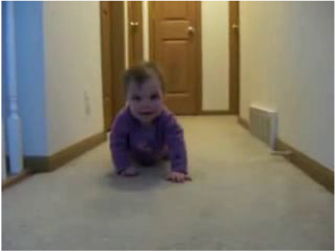}&
\includegraphics[width=2.5cm]{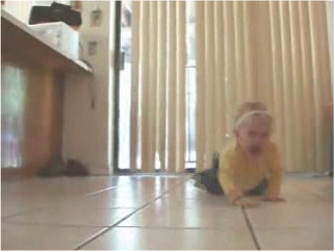}&
\includegraphics[width=2.5cm]{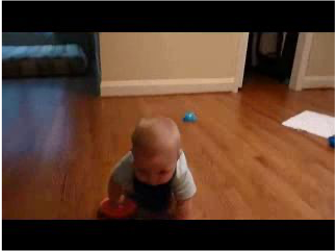}&
\includegraphics[width=2.5cm]{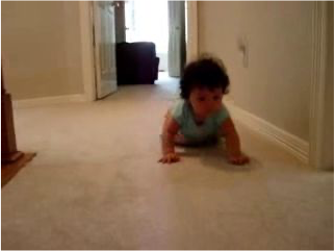}\\
BaseballPitch & 
\includegraphics[width=2.5cm]{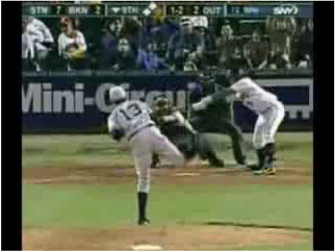}&
\includegraphics[width=2.5cm]{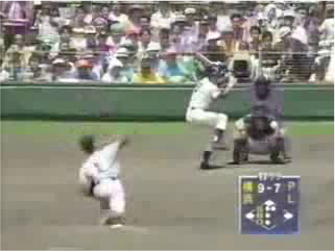}&
\includegraphics[width=2.5cm]{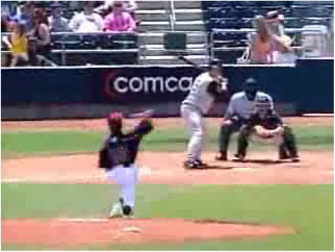}&
\includegraphics[width=2.5cm]{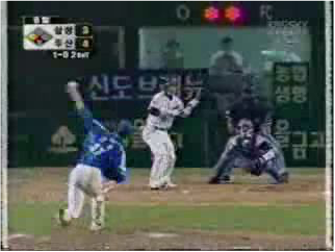}\\
GolfSwing & 
\includegraphics[width=2.5cm]{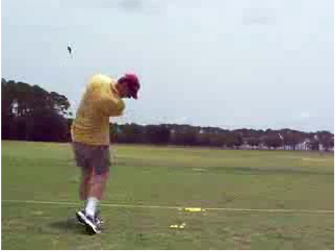}&
\includegraphics[width=2.5cm]{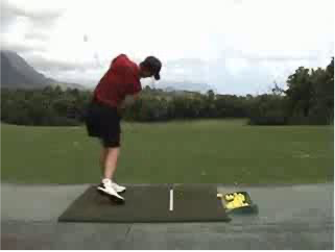}&
\includegraphics[width=2.5cm]{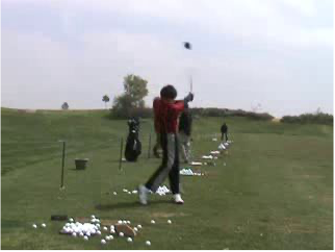}&
\includegraphics[width=2.5cm]{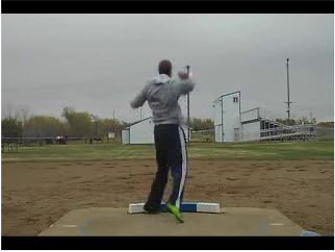}\\
HorseRace & 
\includegraphics[width=2.5cm]{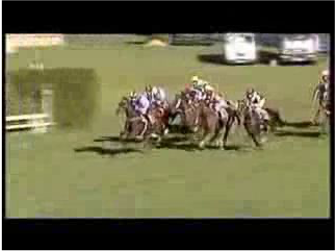}&
\includegraphics[width=2.5cm]{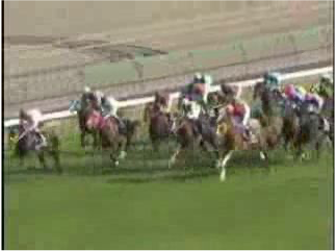}&
\includegraphics[width=2.5cm]{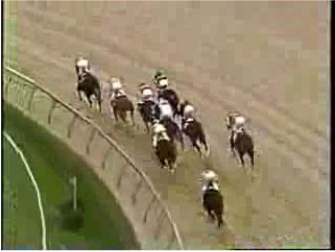}&
\includegraphics[width=2.5cm]{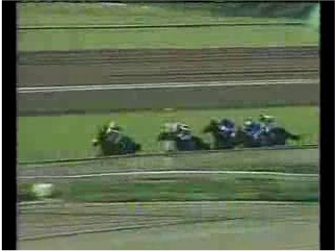}\\
IceDancing & 
\includegraphics[width=2.5cm]{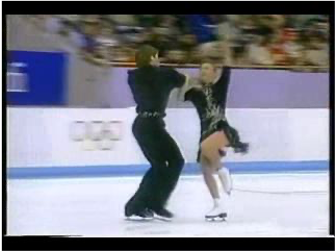}&
\includegraphics[width=2.5cm]{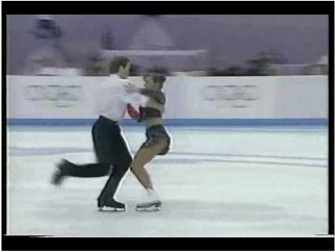}&
\includegraphics[width=2.5cm]{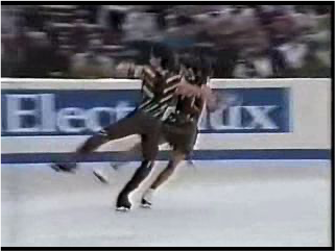}&
\includegraphics[width=2.5cm]{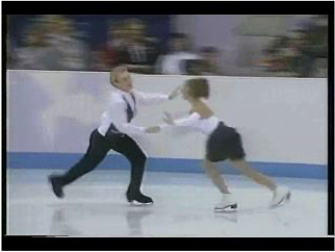}\\
HighJump & 
\includegraphics[width=2.5cm]{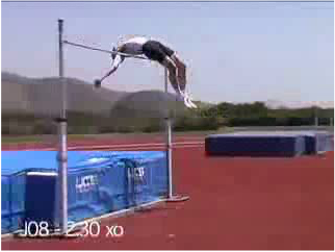}&
\includegraphics[width=2.5cm]{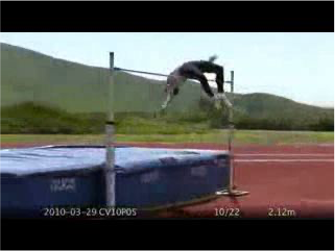}&
\includegraphics[width=2.5cm]{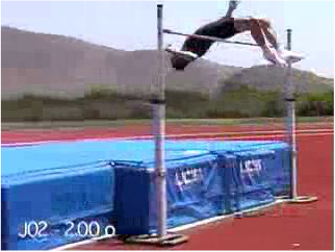}&
\includegraphics[width=2.5cm]{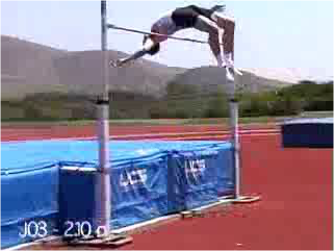}\\
BalanceBeam & 
\includegraphics[width=2.5cm]{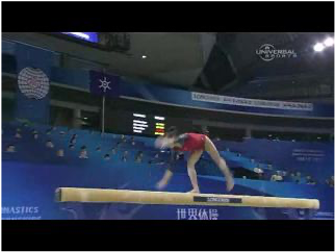}&
\includegraphics[width=2.5cm]{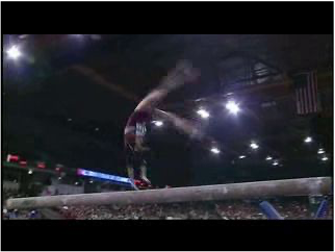}&
\includegraphics[width=2.5cm]{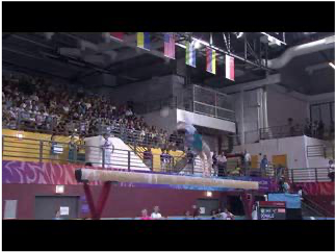}&
\includegraphics[width=2.5cm]{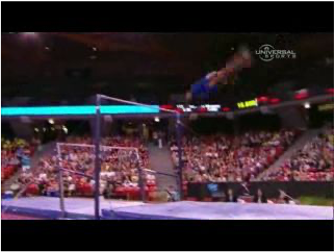}\\
ParallelBars & 
\includegraphics[width=2.5cm]{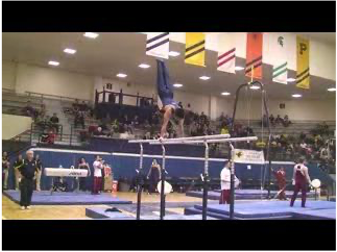}&
\includegraphics[width=2.5cm]{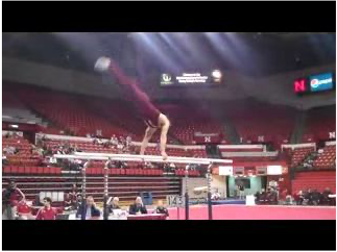}&
\includegraphics[width=2.5cm]{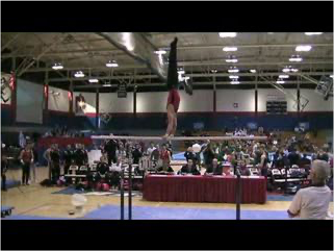}&
\includegraphics[width=2.5cm]{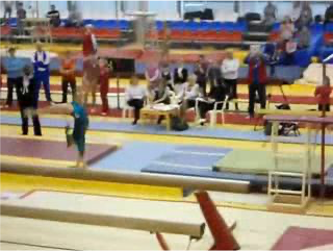}
\end{tabular}
\caption{Video retrieval visualization. The first and second column are categories and images of test instance, respectively. The rightmost 3 columns are nearest retrieval results. We select an image from the video for demonstration.}
\label{fig:videoretrieval} 
\end{figure}

In Figure \ref{fig:videoretrieval}, we visualize the video retrieval result of the integrated pretrained model (SimCLR+SR+TC). We pretrain the model on Kinetics-400 and conduct retrieval experiment on UCF101. It can be observed that our retrieval results are not perfect in that videos from different categories can be retrieved if having similar actions and contents, e.g. image 3 (ApplyLipstick) of ApplyEyeMakeup, image 3 (Shotput) of GolfSwing, image 3 (ParallelBars) of BalanceBeam and image 3 (BalanceBeam) of ParallelBars.

\bibliography{egbib}